\documentclass{article}

     \usepackage[preprint,nonatbib]{neurips_2024}

\usepackage[utf8]{inputenc} 
\usepackage[T1]{fontenc}    
\usepackage{hyperref}       
\usepackage{url}            
\usepackage{booktabs}       
\usepackage{amsfonts}       
\usepackage{nicefrac}       
\usepackage{microtype}      
\usepackage{xcolor}
\usepackage{graphicx}
\usepackage{amsmath}
\usepackage{mathtools}         
\usepackage{multirow}

\title{Efficient Image Deblurring Networks based on Diffusion Models}

    \author{
        Kang Chen, Yuanjie Liu\thanks {Corresponding author.} \\
        College of Information and Electrical Engineering\\
        China Agricultural University \\
    }

\begin{document}

    \maketitle

    \begin{abstract}
        This article presents a sliding window model for defocus deblurring, named Swintormer, which achieves the best performance to date with remarkably low memory usage.
        This method utilizes a diffusion model to generate latent prior features, aiding in the restoration of more detailed images.
        Additionally, by adapting the sliding window strategy, it incorporates specialized Transformer blocks to enhance inference efficiency.
        The adoption of this new approach has led to a substantial reduction in Multiply-Accumulate Operations (MACs) per iteration, drastically cutting down memory requirements.
        In comparison to the currently leading GRL method, our Swintormer model significantly reduces the computational load that must depend on memory capacity, from 140.35 GMACs to 8.02 GMACs, while improving the Peak Signal-to-Noise Ratio (PSNR) for defocus deblurring from 27.04 dB to 27.07 dB.
        This innovative technique enables the processing of higher resolution images on memory-limited devices, vastly broadening potential application scenarios.
        The article wraps up with an ablation study, offering a comprehensive examination of how each network module contributes to the final performance.
    \end{abstract}

    \section{Introduction}

    Image deblurring is a classic task in low-level computer vision, which aims to restore the image from a degraded input and has a wide range of application scenarios.
    Existing networks based on supervised deep learning regression methods such as Restormer~\cite{Zamir2021Restormer}, GRL~\cite{li2023grl} show strong capabilities for image deblurring tasks.
    However, such supervised algorithms invariably demand a considerable volume of labeled data to effectively train regression models.
    Data annotation is labor-intensive and often necessitates domain expertise, thereby resulting in elevated costs.
    Conversely, unsupervised learning methods~\cite{8579082, rombach2021highresolution, DBLP:conf/nips/HoJA20} obviate the need for labeled data, rendering them particularly well-suited for large-scale datasets.
    Nevertheless, acquiring extensive data can pose challenges in certain tasks, and in such contexts, unsupervised methods may not prove to be as efficacious as supervised algorithms.
    Another important issue is the generalization ability.
    The current deblurring algorithms~\cite{abuolaim2020defocus,son2021single_kpac,lee2021iterative,wang2021uformer} often suffer severe performance degradation when confronted with varying data distributions across different scenarios.

    In recent years, attention mechanisms~\cite{7274732, abuolaim2020defocus,lee2021iterative,son2021single_kpac} have demonstrated their effectiveness in enhancing the fitting capability of deep learning regression models.
    Convolution operations, which are often used to implement attention mechanisms, provide local connectivity and translation invariance.
    While convolution operations enhance efficiency and generalization, the convolution operator's receptive field is restricted, impeding its ability to model long-range pixel dependencies.
    As a result, Transformer-based algorithms utilizing the self-attention (SA) mechanism ~\cite{vaswani2017attention, wang2018non,zhang2019self_sagan, vision_transformer} were introduced to address this challenge.
    Although SA is highly effective in capturing long-range pixel interactions, its complexity grows quadratically with spatial resolution.
    This makes it impractical to apply to high resolution images, which is often the case in image deblurring.
    Recently, there have been some attempts to customize the attention mechanism for image deblurring tasks ~\cite{liang2021swinir,wang2021uformer,Zamir2021Restormer}.
    To reduce the computational loads, these methods either apply SA on small $8\times8$ spatial windows around each pixel ~\cite{liang2021swinir,wang2021uformer,li2023grl}, or substitute the fully connected layer in the attention mechanism with a more
    efficient convolution operation ~\cite{Zamir2021Restormer} with sparsity.

    In addition to regression-based methods, leveraging deep generative models presents a promising approach for enhancing image deblurring tasks.
    Noteworthy deep generative models such as Generative Adversarial Networks (GANs)~\cite{goodfellow2014generative,Karras2019stylegan2}, Variational Auto-Encoding model~\cite{kingma2022autoencoding} and Diffusion models (DM)~\cite{DBLP:conf/nips/HoJA20} have demonstrated remarkable performance in image synthesis and super-resolution tasks.
    Moreover, these generative models are typically trained using unsupervised algorithms, obviating the need for labeled data.
    In contrast, many existing supervised tasks rely on image datasets containing only a few hundred images, thereby providing limited image information.
    However, generative models exhibit the ability to learn intricate distributions from extensive datasets, such as ImageNet~\cite{russakovsky2015imagenet}, CIFAR~\cite{swofford2018image}, and COCO~\cite{lin2015microsoft}, which extends beyond the constraints of current labeled datasets.
    Nevertheless, it's important to note that generative models, especially state-of-the-art Diffusion models (DM), often necessitate more computational resources compared to regression models.
    With billions of parameters, these models significantly increase computational overhead.

    To address this limitation, we explored the subtle modeling capability of DM by directly using Robin's pre-trained super-resolution DM~\cite{rombach2021highresolution} to produce high-resolution images on the deblurring dataset DPDD~\cite{abuolaim2020defocus}.
    Remarkably, even after the inevitable downsampling to match the input image size, as shown in Table~\ref{table:Motivation}, we observe a substantial performance enhancement of 0.24dB for single pixel defocus deblurring.
    This attests to DM’s proficiency in deblurring tasks.
    DM’s ability of implicit modelling benefits from a surfeit of training data with diverse images, improving its understanding of image content and structure for detail recovery in various conditions.
    DM uses maximum likelihood estimation to find parameters best fitting the training data distribution:
    \begin{equation}
        \nabla_\theta\Vert \epsilon -\epsilon_\theta(z_{t},t) \Vert_{2}^{2} \,.
        \label{eq:dmloss}
    \end{equation}
    where $z$ represents prior features, $t \in [1, T]$ is a random time-step and $\epsilon \sim \mathcal{N}(\mathbf{0}, \mathbf{I})$ denotes sampled noise.
    The unique training strategy has given DM a significant advantage in its ability to generalize across various conditions.
    However, it is not well-suited to supervised learning tasks like classification or regression, as DM's main focus is on modeling the data distribution rather than predicting specific target values.
    Consequently, we are now undertaking an exploration of fine-tuning and adjustments rooted in the DM paradigm, in a collaborative effort to improve its performance across a range of tasks.

    In addition to the aforementioned challenges, another often overlooked issue is the inconsistency between training and inference.
    Take Restormer~\cite{Zamir2021Restormer} for example, where the training process involves a tensor input of $128\times128\times3\times8$, yet during inference, a $1680\times1120\times3\times1$ tensor is used.
    This discrepancy between training and inference can detrimentally affect a model's performance.
    A recent approach, TLC~\cite{chu2021revisiting}, addresses this issue by substituting global operations (like global average pooling) with local operations at inference time.
    However, this module-swapping technique doesn't universally apply to all models.
    To tackle this challenge, we propose a sliding window-based approach aimed at addressing this inconsistency.

    Based on the above analysis, we propose a novel approach for image deblurring that is capable of fusing more image information and also applicable for large images.
    Firstly, we use a DM model that loads the best pre-trained parameters to generate latent image features based on existing supervised dataset.
    We then proceed to train the deblurring network by inputting the latent image features and the label dataset simultaneously.
    Lastly, we present a broader approach to addressing the inconsistency in the distribution of global information during training and inference.
    The main contributions of this work are summarized below:

    \begin{itemize}
        \item We propose Swintormer, a Sliding Window Transformer for multi-scale representation learning on high-resolution images. It incorporates latent image feature fusion to effectively deblur images.

        \item We propose a novel training scheme that utilizes DM to generate image features conditionally, enabling the model to learn additional prior information.
        \item We propose a more efficient method for attention computation in image processing, which includes both channel attention and spatial attention. The channel attention is a multi-Dconv head transposed attention and the spatial attention is a shifted windows-Dconv attention (SWDA).
        \item We present a more general approach by dividing the image into overlapping patches for independent inference, which improves the model performance.
    \end{itemize}


    \section{Related Works}

    \subsection{Image deblurring}
    The traditional deblurring algorithm typically involves formulating and solving an optimization problem based on the causes of image blurring~\cite{karaali2017edge_EBDB,shi2015just_jnb,10.1145/1179352.1141956,8048543}.
    However, these approaches depend on manually designed image features, leading to limited generalization capability and constrained performance in intricate scenarios.
    Currently, deep learning-based on image deblurring focuses on establishing a direct mapping between blurred images and sharp images from paired datasets:
    \begin{equation}
        I_b=\phi(I_s;\theta_i),
    \end{equation}
    where $\phi$ is the image blur function, and $\theta_i$ is a parameter vector. $I_s$ is the sharp image. $I_b$ is the blurred image.
    With the powerful fitting capability of deep learning, it is possible to directly train the model end-to-end to learn this mapping, thereby achieving deblurring~\cite{wang2021uformer,li2023grl,Zamir2021Restormer}.
    Current research primarily focuses on general algorithms that aim to improve model representation by using advanced neural network architecture designs such as residual blocks, dense blocks, attention blocks, and others~\cite{
        wang2021learning, liang2021fkp, liang21hcflow, liang21manet, guo2020closed, cheng2021mfagan, deng2021deep, fu2019jpeg, kim2019pseudo, fu2021model}.
    Notably, the Transformer architecture~\cite{vaswani2017attention, vision_transformer} has exhibited remarkable success.
    Numerous experiments have shown that the effectiveness of the Transformer primarily lies in the design of token mixer and the FFN(Feed-Forward Network)~\cite{yu2022metaformer}.
    In particular, the self-attention mechanism in the token mixer is recognized as the key driver of its superior performance.
    However, its complexity increases quadratically as the number of patches grows, making it infeasible for high-resolution images.
    To address this issue, various revised token mixers~\cite{DBLP:journals/corr/abs-1904-10509,2020arXiv200405150B,NEURIPS2020_c8512d14,tay2020sparse,Kitaev2020Reformer,
        2020arXiv200604768W,tay2020synthesizer,katharopoulos2020transformers,choromanski2020rethinking} were developed to reduce complexity in different image processing applications.
    On the other side, different FFN designs such as Mlp~\cite{MLP}, GluMlp~\cite{dauphin2017language_gating}, GatedMlp~\cite{liu2021pay}, ConvMlp~\cite{li2021convmlp} and SimpleGate~\cite{chen2022simple} were proposed.
    While these designs have their own advantages and disadvantages in various low-level visual tasks, the performance difference remains consistent when the number of model parameters is nearly the same.

    In addition to developing general algorithms, another focus is on creating specialized models optimized for specific blurry situations, including image super-resolution (SR), real-world image deblurring, image denoising, and the reduction of JPEG compression artifacts~\cite{ADL2022, Zhang2023kbnet, liang2021swinir, chen2023hat,kawar2022jpeg}.

    \subsection{Diffusion Model}
    Recently, Diffusion Models have emerged as leaders in unconditional image synthesis, leveraging unsupervised learning algorithms to extract priors from datasets and achieve state-of-the-art results.
    In contrast to previous, models such as feed-forward, GAN, and flow-based models, which directly learn a mapping of $f$ from the input $x$ to the result $y$:
    \begin{equation}
        y=f(x),
    \end{equation}
    The diffusion model adopts a distinct approach. It treats the generation process as an optimization computation, expressed as:
    \begin{equation}
        y=\arg \min_y E_\theta(x,y),
    \end{equation}
    where $E$ is the expectation and $\theta$ represents the parameters.
    The information it directly learns is not the joint distribution of pixels, but rather the gradient of the distribution.
    In other words, instead of learning a map directly, DM builds a neural network to find a solution to the optimization problem and then samples the solution to get an image.

    Drawing upon its robust ability to grasp image priors from datasets, various impressive diffusion frameworks have been utilized for low-level vision tasks~\cite{luo2023image,peebles2023scalable,kawar2022denoising,DBLP:conf/miccai/RonnebergerFB15,
        DBLP:conf/nips/HoJA20, DBLP:journals/corr/abs-2011-13456, DBLP:journals/corr/abs-2105-05233,rombach2021highresolution,esser2020taming}.
    Nonetheless, training a highly advanced diffusion model often demands expensive computing resources.
    To address this limitation, we propose to circumvent this drawback with our Swintormer approach.



    \begin{table}
        \centering
        \caption{  Quantitative comparison by applying diffusion model~\cite{rombach2021highresolution} on the DPDD Dataset~\cite{abuolaim2020defocus}.
        In order to make a fair comparison with other existing methods, the input image is an 8-bit image instead of a 16-bit image.}
        \label{table:Motivation}
        \resizebox{0.6\linewidth}{!}{
            \begin{tabular}{c|c|c|c|c}
                \toprule
                Model                            & PSNR~$\textcolor{black}{\uparrow}$ & SSIM~$\textcolor{black}{\uparrow}$ & MAE~$\textcolor{black}{\downarrow}$ & LPIPS~$\textcolor{black}{\downarrow}$ \\
                \hline
                Input                            & 23.70                              & 0.713                              & 0.048                               & 0.354                                 \\
                \hline
                JNB~\cite{shi2015just_jnb}       & 23.69                              & 0.707                              & 0.048                               & 0.442                                 \\
                EBDB~\cite{karaali2017edge_EBDB} & 23.94                              & 0.723                              & 0.047                               & 0.402                                 \\
                DMENet~\cite{lee2019deep_dmenet} & 23.90                              & 0.720                              & 0.047                               & 0.410                                 \\
                \hline
                Diffusion                        & 23.94                              & 0.727                              & 0.047                               & 0.349                                 \\
                \bottomrule
            \end{tabular}
        }
    \end{table}

    \section{Method}
\begin{figure}[t]
        \includegraphics[width=\linewidth]{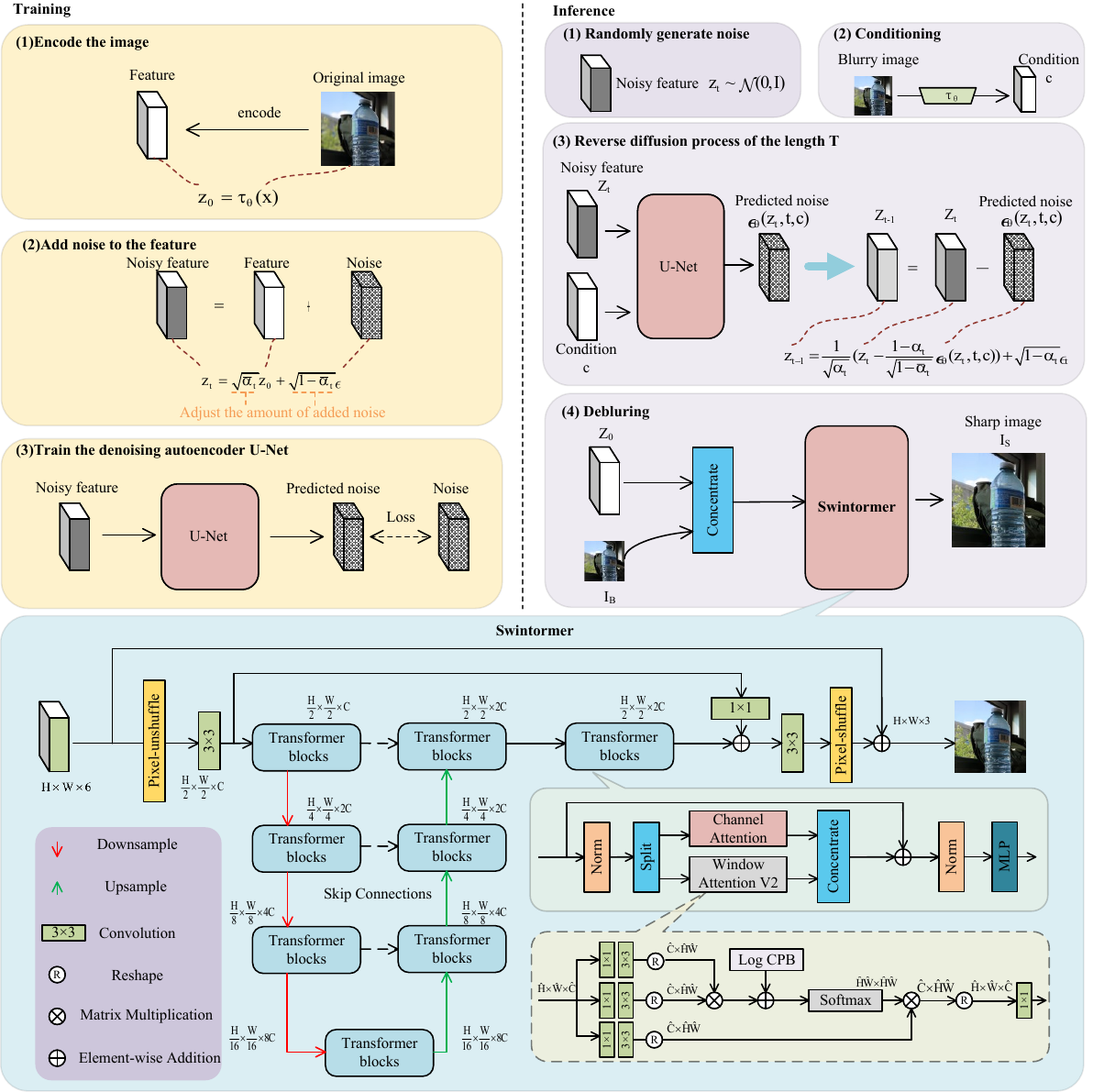}
        \caption{The architecture of the Swintormer.}
        \label{fig:Swintormer}
    \end{figure}

    Our goal is to develop an image deblurring model that efficiently utilizes priors generated by the diffusion model while maintaining the advantage of being memory-efficient.
    We propose Sliding Window Image Restoration Model (Swintormer).
    The Swintormer is a regression-based model designed with a sliding window strategy.
    An overview of the pipeline is presented in Fig.~\ref{fig:Swintormer}.
    In this section, we introduce the design of the transform module in the model and demonstrate how diffusion is employed to generate prior features.

    \subsection{Shifted Windows-Dconv Attention}
    While self-attention~\cite{vaswani2017attention,vision_transformer} is highly effective, the time and memory complexity of the key-query dot-product interaction grows quadratically in self-attention as the input resolution increases.
    For example, performing a calculation on a tensor of size $8\times128\times128\times48$ requires 64GB of video memory.
    Similar to deal with long sentence problems in NLP~\cite{DBLP:journals/corr/abs-1904-10509,2020arXiv200405150B,NEURIPS2020_c8512d14,tay2020sparse,Kitaev2020Reformer,
        2020arXiv200604768W,tay2020synthesizer,katharopoulos2020transformers,choromanski2020rethinking}, many methods~\cite{liang21swinir, Zamir2021Restormer,chen2022simple} are proposed for high-resolution image.
    However, these methods have different performance in different low-level visual tasks.
    Therefore, we propose Swintormer for improving generalization performance, that has linear complexity.
    The key innovation is to segment the feature tensor along the channel dimension and then calculate channel attention and spatial attention separately.
    Another crucial aspect is the use of depth-wise convolutions to generate \emph{query} (\textbf{Q}), \emph{key} (\textbf{K}) and \emph{value} (\textbf{V}) projections instead of using linear layers, which can highlight the local context to accelerate model convergence.

    Given a layer normalized tensor ~$\mathbf{Y}$~$\in$~$\mathbb{R}^{\hat{H}\times \hat{W} \times \hat{C}}$, our SWDA first applies a shifted window partitioning approach to divide $\mathbf{Y}$ into ${M} \times {M} $ patches (with a default window size of 16).
    One challenge with this routine is that some windows may end up smaller than ${M}\times{M}$.
    Therefore, a cyclic-shifting toward the top-left direction method is used to solve the problem~\cite{liu2021Swin}.
    The resulting patches are then used to generate \textbf{Q}, \textbf{K} and \textbf{V} projections through 1×1 convolutions to aggregate pixel-wise cross-channel context, followed by 3×3 depth-wise bias-free convolutions.
    As a result, the tensors \textbf{Q}, \textbf{K} and \textbf{V} are all the same size,$\mathbb{R}^{{M}\times {M} \times \hat{C}}$.
    These tensors are then split into two parts along the channel, each with a size of $\mathbb{R}^{{M}\times {M} \times \frac{C}{2}}$.
    One part is used for channel attention calculation by MDTA~\cite{Zamir2021Restormer}, where their dot-product interaction produces a transposed-attention map of size $\mathbb{R}^{\frac{C}{2}\times \frac{C}{2}}$, while the other part is utilized for spatial attention calculation, resulting in an attention map with size of $\mathbb{R}^{{M}\times {M}}$.
    Overall, the process is defined as:
        {\small
    \begin{equation}
        \text{Channel Attention} = \text{SoftMax}(Q_cK_c^T)V_c,
    \end{equation}
    \begin{equation}
        \text{Spacial Attention} = \text{SoftMax}(Q_sK_s^T+B)V_s,
    \end{equation}
    \begin{equation}
        \text{Attention} = W^{(\cdot)}_p\text{concat(Channel Attention, Spacial Attention)},
    \end{equation}
    }where $Q_c,K_c,V_c$ $\in$ $\mathbb{R}^{\frac{C}{2}\times {M^2}}$;$Q_s,K_s,V_s$ $\in$ $\mathbb{R}^{ {M^2} \times \frac{C}{2}}$;
    $B$ $\in$ $\mathbb{R}^{{M^2}\times {M^2}}$ represents the relative position bias term for each head;$W^{(\cdot)}_p$ denotes the 1×1 point-wise convolution, and $M^2$ is the number of patches in a window.
    The relative position bias encodes the relative spatial configurations of visual elements.

    \subsection{Diffusion Model}
    Many methods, such as Masked Autoencoders~\cite{he2021masked}, improve the performance of the model by destroying the feature and then rebuilding the feature.
    Based on this idea of the feature destruction and reconstruction, we introduce the diffusion model to process some important features.
    Specifically, our diffusion model is based on latent conditional denoising diffusion probabilistic models~\cite{rombach2021highresolution, DBLP:conf/nips/HoJA20}.
    It consists of a forward diffusion process $q(z_{1:T}|z_0)$ and a reverse denoising process $p_\theta(z_{0:T})$ where $T$ is a fixed Markov Chain of length.
    When given a feature $x$, we use VQGAN~\cite{esser2020taming} to obtain the latent space $z=\mathcal{E}(x)$.
    With the latent space, we compute the forward diffusion process for training the diffusion model and calculate the reverse denoising process for generating the processed feature.

    \noindent\textbf{Forward diffusion process.} In the forward process, we input $z$ and gradually add Gaussian noise $\mathcal{N}$ to it according to a variance schedule $\beta_1, \dotsc, \beta_T$:
    \begin{equation}
        q(z_{1:T} | z_0) \coloneqq \prod_{t=1}^T q(z_t | z_{t-1} ), \qquad
    \end{equation}
    \begin{equation}
        q(z_t|z_{t-1}) \coloneqq \mathcal{N}(z_t;\sqrt{1-\beta_t}z_{t-1},\beta_t \mathbf{I})
        \label{eq:forwardprocess}
    \end{equation}
    By fixing the variances $\beta_t$ to constants and the reparameterization~\cite{kingma2013auto} $z_t(z_0, {\epsilon}) = \sqrt{\bar\alpha_t}z_0 + \sqrt{1-\bar\alpha_t}{\epsilon}$,
    for ${\epsilon} \sim \mathcal{N}(\mathbf{0}, \mathbf{I})$,
    thus Eq.~\eqref{eq:forwardprocess} can be rewritten as:
    \begin{equation}
        q(z_t|z_0) = \mathcal{N}(z_t; \sqrt{\bar\alpha_t}z_0, (1-\bar\alpha_t)\mathbf{I})
        \label{eq:q_marginal_arbitrary_t}
    \end{equation}
    where $\alpha_t \coloneqq 1-\beta_t$ and $\bar\alpha_t \coloneqq \prod_{i=1}^t \alpha_i$.
    Specifically, we trained a denoising network $\boldsymbol\epsilon_\theta$ by predicting ${\epsilon}$ from $z_t$.

    \noindent\textbf{Reverse denoising process.} In the reverse process $p_\theta(z_{0:T})$:
    \begin{equation}
    {p_{\theta}}(z_{0:T})
        \coloneqq p(z_T)\prod_{t=1}^T p_\theta(z_{t-1}|z_t), \qquad
    \end{equation}
    \begin{equation}
        p_\theta(z_{t-1}|z_t) \coloneqq \mathcal{N}(z_{t-1};{\boldsymbol{\mu}}_\theta(z_t, t), {\boldsymbol{\Sigma}}_
        \theta(z_t, t))
    \end{equation}

    Following previous work~\cite{DBLP:conf/nips/HoJA20} to accelerate reverse processing by reparameterization, we
    generated the prior feature $z_0$ using KL divergence to directly compare $p_\theta(z_{0:T})$ against forward process posteriors from $z_t$ to $z_0$:

    \begin{equation}
        q(z_{t-1}|z_t,z_0) =  \mathcal{N}(z_{t-1}; \tilde{\boldsymbol{\mu}}_t(z_t, z_0), \tilde\beta_t \mathbf{I}),
        \label{eq:q_posterior_mean_var}
    \end{equation}

    \begin{equation}
        \text{where} \qquad \tilde{\boldsymbol{\mu}}_t(z_t, z_0) \coloneqq \frac{\sqrt{\bar\alpha_{t-1}}\beta_t }{1-
        \bar\alpha_t}z_0 + \frac{\sqrt{\alpha_t}(1- \bar\alpha_{t-1})}{1-\bar\alpha_t} z_t,
    \end{equation}

    \begin{equation}
        \text{and} \qquad \tilde\beta_t \coloneqq \frac{1-\bar\alpha_{t-1}}{1-\bar\alpha_t}\beta_t
    \end{equation}

    Consequently, with the trained denoising network $\boldsymbol \epsilon_\theta$  conditioned on the $c$ (The condition $c$ is usually images~\cite{Park_2019_CVPR},but can also be text and semantic
    maps~\cite{Isola2017ImagetoImageTW, Reed2016GenerativeAT}) for predicting the noise ${\epsilon}$, we can iteratively sample $z_t$ as follows:
    \begin{equation}
        z_{t-1} = \frac{1}{\sqrt{\alpha_t}}( z_t - \frac{1-\alpha_t}{\sqrt{1-\bar\alpha_t}} {\boldsymbol{\epsilon}}_\theta(z_t, t, c)) + \sqrt{1-\alpha_t} {{\epsilon_t}}
        \label{eq:ddpm}
    \end{equation}
    where ${\epsilon_t} \sim \mathcal{N}(\mathbf{0}, \mathbf{I})$.
    After T iterations, we can get feature $z_0$ as illustrated in Fig.~\ref{fig:Swintormer}.
    We further explore the iteration numbers T in Section~\ref{Ablation}.


    \subsection{Inference Strategy}
    We introduce a novel strategy to ensure consistency between the input tensor sizes used in both training and inference.
    This is achieved by incorporating a pre-processing operation that employs shifted windows before the inference step.
    While this pre-processing operation may introduce additional computational overhead as overlapping regions are redundantly processed by the entire model,
    it also increases parallelization due to maintaining the same batch size, thereby expediting the inference process.
    Moreover, this approach grants control over the size of the overlapping region by adjusting the sliding pace, allowing for a more nuanced trade-off between deblurring quality and inference speed in practical applications.

    \subsection{Training Strategy}  \label{Training Strategy}
    First, it is aimed to establish a diffusion model for feature extraction, and then train Swintormer for deblurring.
    We utilize a super-resolution LDM~\cite{DBLP:journals/corr/abs-2105-05233} to execute the diffusion model.
    This specific LDM is chosen because its generated features align more closely with the input image's own feature distribution, rather than integrating the overall feature distribution of other images in the training dataset.
    The super-resolution LDM consists of a denoising autoencoder $\boldsymbol\epsilon_\theta$ and a VQGAN model $\tau_\theta$.
    The VQGAN is a pre-trained model, and in our approach, regardless of the training or inference stage, its parameters are frozen, and we only need to train the denoising autoencoder $\boldsymbol\epsilon_\theta$ in the LDM.
    $\boldsymbol\epsilon_\theta$ is a time-conditioned U-Net denoising autoencoder~\cite{rombach2021highresolution, DBLP:conf/miccai/RonnebergerFB15}.
    Specifically, we first use $\tau_\theta(x)$ to encode the input image and obtain the feature $z_0$.
    Then in the latent space, we iteratively add Gaussian noise to the input feature $z_0$ to obtain the blurred feature $z_t$.
    During this diffusion process, we train the denoising autoencoder $ \epsilon_\theta(z_{t},t,c);\, t=1\dots T$  to make its estimated noise consistent with the Gaussian noise we introduce:
    \begin{equation}
        L_{LDM} \coloneqq \mathbb{E}_{\mathcal{E}(x),y, \epsilon \sim \mathcal{N}(0, I), t }\Big[ \Vert \epsilon -
        \epsilon_\theta(z_{t},t, \tau_\theta(y)) \Vert_{2}^{2}\Big] \, ,
        \label{eq:cond_loss}
    \end{equation}
    Here, the input $y$ is the blurred image used to guide the diffusion process.

    After that, we proceed to train the Swintormer deblurring model.
    We utilize the trained denoising autoencoder model $\boldsymbol\epsilon_\theta$ to estimate the noise present in the blurred image.
    The estimated noise is then used through Eq.~(\ref{eq:ddpm}) to sample and acquire the prior feature $z_0$.
    Subsequently, along with the corresponding blurred image, it is employed in the training of the Swintormer model $\phi_\theta$.
    This training process incorporates the utilization of L$_1$ loss and perceptual loss as Eq.~(\ref{eq:l1}) and Eq.~(\ref{eq:ploss}):
    \begin{equation}
        \label{eq:l1}
        \mathcal{L}_{deblur}=\Vert I_s-\phi_\theta(I_b, z_0)\Vert_1
    \end{equation}
    \begin{equation}
        \label{eq:ploss}
        \mathcal{L}_{deblur}=\Vert VGG(I_s)-VGG(\phi_\theta(I_b, z_0))\Vert_2^2
    \end{equation}
    Here, $\phi_\theta$ is the deblurring model Swintormer, and VGG is the associated feature extraction model~\cite{simonyan2014very} during the deblurring process.
    We will explain the performance difference between these two loss functions in Section~\ref{Experiments}.

    \subsection{Inference} \label{Inference}
    Using the trained denoising autoencoder and Swintormer for deblurring involves two corresponding stages.
    First, prior feature extraction is performed.
    The blurred image $x$~$\in$~$\mathbb{R}^{H\times W \times 3}$ to be processed is fed into the denoising autoencoder $\boldsymbol\epsilon_\theta$, and the resulting encoded result $z_t$ is diffused through the DDIM~\cite{song2020denoising} to obtain the prior feature $z_0$~$\in$~$\mathbb{R}^{H\times W \times 3}$.
    It is worth noting that that the prior feature $z_0$ will not be decoded by the VQGAN model $\tau_\theta$.
    Instead, along the channel dimension, $z_0$ is concatenated with x to form an extended feature tensor $x_f$~$\in$~$\mathbb{R}^{H\times W \times 6}$ for deblurring computation.
    To reduce the distribution shifts between training and inference, we partition the feature tensor into the overlapping patches,
    resulting in the input tensor denoted as $x_{input}$~$\in$~$\mathbb{R}^{M\times M \times C \times B}$.
    Here, $M$ signifies the window size, $B$ is the training batch size, and the dimensions of the input tensor.
    After processing by Swintormer, deblurred image patches $x_{dp}$~$\in$~$\mathbb{R}^{M\times M \times 3 \times B}$ are obtained.
    Finally, the resulting patches are merged into a complete deblurred image $x_{db}$~$\in$~$\mathbb{R}^{H\times W \times 3}$, where the overlapping regions are averaged to generate the corresponding value.

    \section{Experiments and Analysis} \label{Experiments}
    We evaluate the proposed Swintormer on benchmark datasets for two tasks: \textbf{(a)} defocus deblurring, and \textbf{(b)} single-image motion deblurring.
    More details about the datasets, training protocols, and extra visual outcomes are available in the technical appendices.
    In tables showing the performance of the evaluated methods, the top scores are \textbf{highlighted}.

    \begin{table}[h]
        \centering
        \caption{Single-image defocus deblurring results on the RealDOF dataset~\cite{lee2021iterative}.
        Our method outperforms existing baselines without extra training data.}
        \label{table:realdeblurring}
        \resizebox{0.5\linewidth}{!}{
            \begin{tabular}{c|c|c|c}
                \hline
                Method                              & PSNR~$\textcolor{black}{\uparrow}$ & SSIM~$\textcolor{black}{\uparrow}$ & LPIPS~$\textcolor{black}{\downarrow}$ \\
                \hline
                DPDNet~\cite{abuolaim2020defocus}   & 22.87                              & 0.670                              & 0.425                                 \\
                KPAC~\cite{son2021single_kpac}      & 23.98                              & 0.716                              & 0.336                                 \\
                IFAN~\cite{lee2021iterative}        & 24.71                              & 0.749                              & 0.306                                 \\
                Restormer~\cite{Zamir2021Restormer} & 25.08                              & 0.769                              & 0.289                                 \\
                DRBNet~\cite{ruan2022learning}      & 25.75                              & 0.771                              & \textbf{0.257}                        \\
                Swintormer(ours)                    & \textbf{25.83}                     & \textbf{0.772}                     & \textbf{0.257}                        \\
                \hline
            \end{tabular}}
    \end{table}

    \begin{figure*}[h]
        \begin{center}
            \resizebox{\linewidth}{!}{
                \begin{tabular}{c@{ } c@{ }   c@{ } c@{ }   }
                    \hspace{-4mm}
                    \multirow{3}{*}[18pt]{\includegraphics[width=0.25\textwidth]{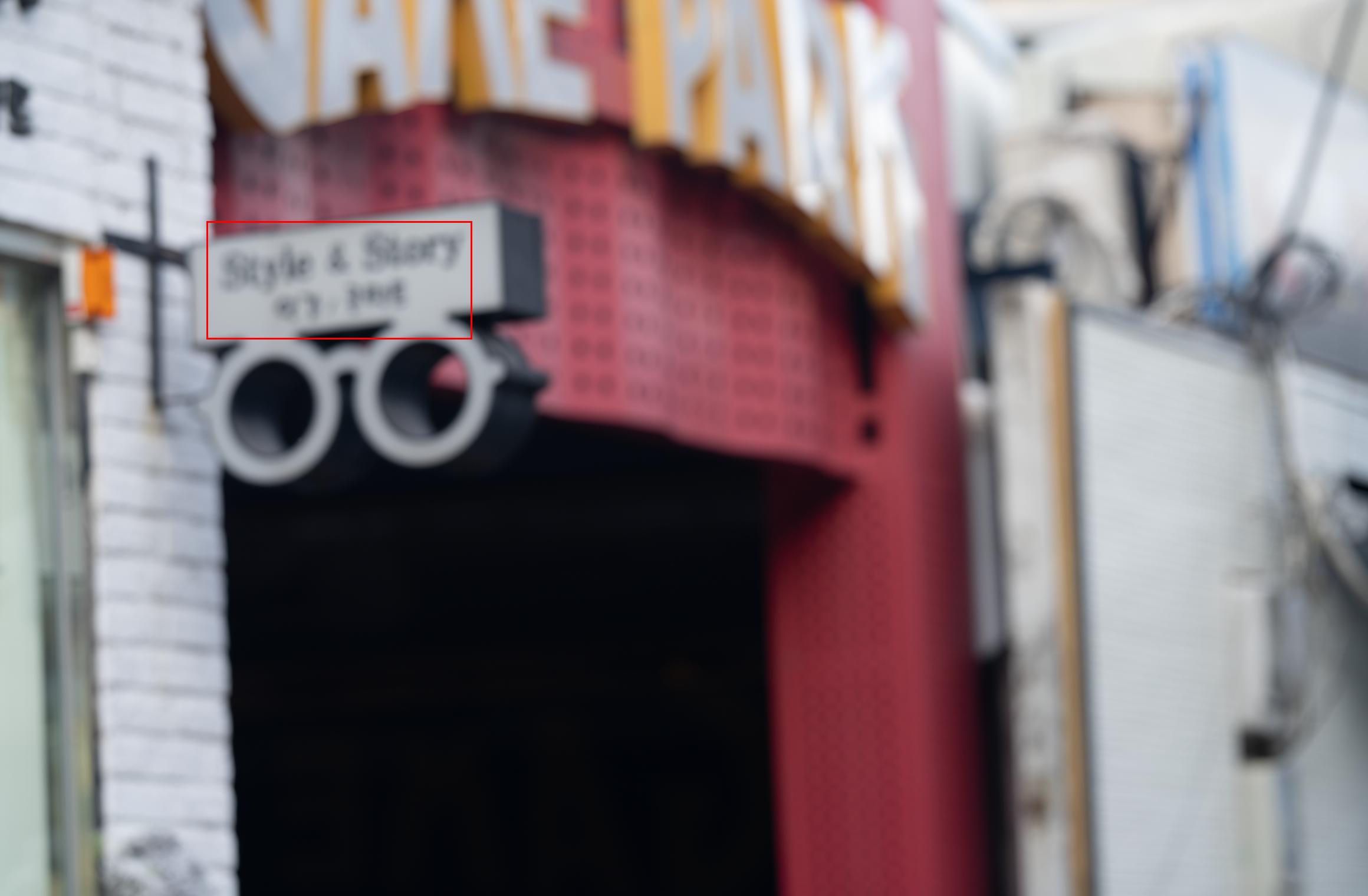}} &
                    \includegraphics[width=.145\textwidth]{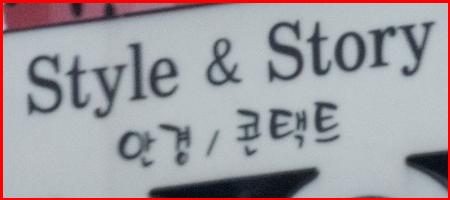} &
                    \includegraphics[width=.145\textwidth]{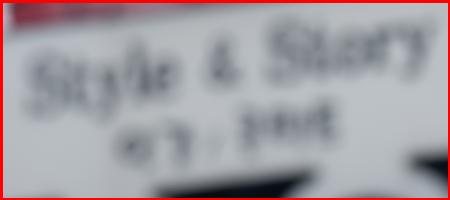} &
                    \includegraphics[width=.145\textwidth]{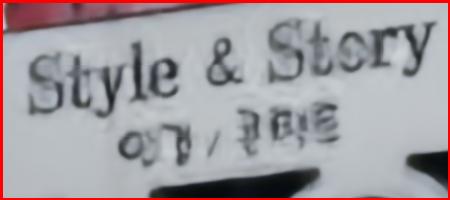}
                    \\
                    & \scriptsize~Reference & \scriptsize~Blurry & \scriptsize~IFAN~\cite{lee2021iterative}
                    \\
                    &
                    \includegraphics[width=.145\textwidth]{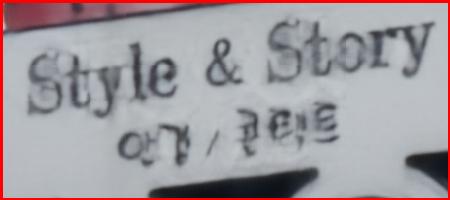} &
                    \includegraphics[width=.145\textwidth]{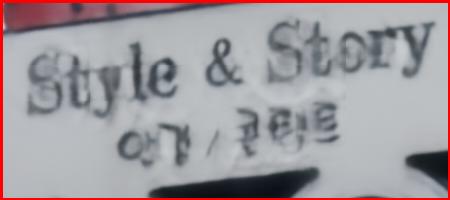} &
                    \includegraphics[width=.145\textwidth]{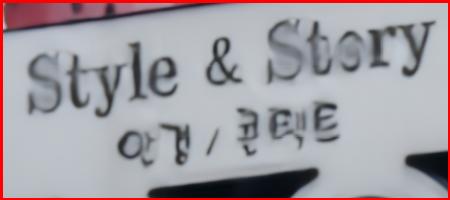}
                    \\
                    \scriptsize~Blurry Image & \scriptsize~Restormer~\cite{Zamir2021Restormer} & \scriptsize~DRBNet~\cite{ruan2022learning} & \scriptsize~\textbf{Ours}
                \end{tabular}
            }
        \end{center}
        \caption{{Visual comparison on the RealDOF dataset~\cite{lee2021iterative}.}}
        \vspace{-1em}
    \end{figure*}

    \subsection{Defocus Deblurring Results}
    As shown in Table~\ref{table:realdeblurring}, our Swintormer performs best on the RealDOF dataset~\cite{lee2021iterative}.
    Table~\ref{table:dpdeblurring} shows image fidelity scores compared with conventional methods.
    Both version of our Swintormer trained under the two different loss functions have advantages compared with state-of-the-art.
    Particularly, in the outdoor scene category, Swintormer yields $~0.12$ dB improvements over the previous best method GRL~\cite{li2023grl}.
    And, it is worth noting that our method achieves the highest perceptual scores in terms of LPIPS on all scene categories.

    \begin{table}[h]
        \caption{Deblurring results on the DPDD dataset(containing 37 indoor and 39 outdoor scenes).
        Swintormer sets new state-of-the-art in metric PSNR and LPIPS by using the L$_1$ loss function and the
        Perceptual loss function respectively.
        \textbf{S:} single-image defocus deblurring. \textbf{D:} dual-pixel defocus deblurring.
        }
        \label{table:dpdeblurring}
        \resizebox{\linewidth}{!}{
            \begin{tabular}{l | c c c c | c c c c | c c c c }
                \toprule[0.15em]
                & \multicolumn{4}{c|}{\textbf{Indoor Scenes}} & \multicolumn{4}{c|}{\textbf{Outdoor Scenes}} & \multicolumn{4}{c}{\textbf{Combined}} \\
                \cline{2-13}
                \textbf{Method}
                & PSNR~$\textcolor{black}{\uparrow}$ & SSIM~$\textcolor{black}{\uparrow}$ & MAE~$\textcolor{black}{\downarrow}$ & LPIPS~$\textcolor{black}{\downarrow}$
                & PSNR~$\textcolor{black}{\uparrow}$ & SSIM~$\textcolor{black}{\uparrow}$& MAE~$\textcolor{black}{\downarrow}$ & LPIPS~$\textcolor{black}{\downarrow}$
                & PSNR~$\textcolor{black}{\uparrow}$ & SSIM~$\textcolor{black}{\uparrow}$& MAE~$\textcolor{black}{\downarrow}$ & LPIPS~$\textcolor{black}{\downarrow}$   \\
                \midrule[0.15em]
                EBDB$_S$~\cite{karaali2017edge_EBDB}      & 25.77          & 0.772          & 0.040          & 0.297           & 21.25          & 0.599          & 0.058          & 0.373 & 23.45 & 0.683 & 0.049 & 0.336 \\
                DMENet$_S$~\cite{lee2019deep_dmenet}      & 25.50          & 0.788          & 0.038          & 0.298           & 21.43          & 0.644          & 0.063          & 0.397 & 23.41 & 0.714 & 0.051 & 0.349 \\
                JNB$_S$~\cite{shi2015just_jnb}            & 26.73          & 0.828          & 0.031          & 0.273           & 21.10          & 0.608          & 0.064          & 0.355           & 23.84 & 0.715 & 0.048 & 0.315 \\
                DPDNet$_S$~\cite{abuolaim2020defocus}     & 26.54          & 0.816          & 0.031          & 0.239           & 22.25          & 0.682          & 0.056          & 0.313 & 24.34 & 0.747 & 0.044 & 0.277\\
                KPAC$_S$~\cite{son2021single_kpac}        & 27.97          & 0.852          & 0.026          & 0.182           & 22.62          & 0.701          & 0.053          & 0.269 & 25.22 & 0.774 & 0.040 & 0.227 \\
                IFAN$_S$~\cite{lee2021iterative}          & {28.11}        & {0.861}        & {0.026}        & {0.179}         & {22.76}        & {0.720} & {0.052}  & {0.254}  & {25.37} & {0.789} & {0.039} & {0.217}\\
                {Restormer}$_S$~\cite{Zamir2021Restormer} & {28.87}        & {0.882}        & {0.025}        & {0.145}         & {23.24}        & {0.743}        & {0.050}        & {0.209}   & {25.98}  & {0.811}  & {0.038}  & {0.178}   \\
                {GRL}$_S$~\cite{li2023grl}                & \textbf{29.06} & \textbf{0.886} & \textbf{0.024} & \textbf{0.139 } & {23.45} & {0.761}  & {0.049} & \textbf{ 0.196}  & \textbf{26.18}  & {0.822 }  & {0.037}  & \textbf{0.168}   \\
                \textbf{Swintormer}$_S$-Perceptual        & {28.95}        & {0.883}        & {0.025}        & {0.141}         & {23.33}        & {0.750}  & {0.050} & {0.205}  & {26.09}  & {0.819}  & {0.038}  & \textbf{0.168}   \\
                \textbf{Swintormer}$_S$-L1                & {28.99}        & {0.884}        & {0.025}        & {0.142}         & \textbf{23.51} & \textbf{0.769}  & \textbf{0.042} & {0.209}  & \textbf{26.18}  & \textbf{0.823}  & \textbf{0.034}  & {0.176}   \\
                \midrule[0.1em]
                \midrule[0.1em]
                DPDNet$_D$~\cite{abuolaim2020defocus}     & 27.48          & 0.849          & 0.029          & 0.189           & 22.90          & 0.726          & 0.052          & 0.255 & 25.13 & 0.786 & 0.041 & 0.223 \\
                RDPD$_D$~\cite{abdullah2021rdpd}          & 28.10          & 0.843          & 0.027          & 0.210           & 22.82          & 0.704          & 0.053          & 0.298 & 25.39 & 0.772 & 0.040 & 0.255 \\
                Uformer$_D$~\cite{wang2021uformer}        & 28.23          & 0.860          & 0.026          & 0.199           & 23.10          & 0.728          & 0.051          & 0.285 & 25.65 & 0.795 & 0.039 & 0.243 \\
                IFAN$_D$~\cite{lee2021iterative}          & {28.66}        & {0.868}        & {0.025}        & {0.172}         & {23.46}        & {0.743} & {0.049} & {0.240} & {25.99} & {0.804} & {0.037} & {0.207} \\
                {Restormer}$_D$~\cite{Zamir2021Restormer} & {29.48}        & {0.895}        & {0.023}        & {0.134}         & {23.97}        & {0.773}        & {0.047}        & {0.175}   & {26.66}  & {0.833}  & {0.035}  & {0.155} \\
                {GRL}$_D$~\cite{li2023grl}                & \textbf{29.83} & \textbf{0.903} & \textbf{0.022} & {0.114}         & {24.39}        & {0.795 }       & \textbf{0.045} & { 0.150 }  & {27.04}  & \textbf{0.847}  & \textbf{0.034}  & {0.133} \\

                \textbf{Swintormer}$_D$-Perceptual        & {29.55}        & {0.897}        & {0.023}        & \textbf{0.107}  & {24.40}  & {0.796}  & \textbf{0.045} & \textbf{0.147}  & {26.91}  & {0.845}  & \textbf{0.034}  & \textbf{0.128} \\
                \textbf{Swintormer}$_D$-L1                & {29.74}        & {0.899}        & \textbf{0.022} & {0.127}         & \textbf{24.52}  & \textbf{0.798}  & \textbf{0.045} & {0.167}  & \textbf{27.07}  & \textbf{0.847}  & \textbf{0.034}  & {0.148} \\
                \bottomrule[0.1em]
            \end{tabular}}
    \end{table}

    \begin{figure*}[h]
        \begin{center}
            \resizebox{\linewidth}{!}{
                \begin{tabular}[b]{c@{ } c@{ }  c@{ } c@{ } c@{ }   }
                    \hspace{-4mm}
                    \multirow{3}{*}[25pt]{\includegraphics[width=0.30\textwidth]{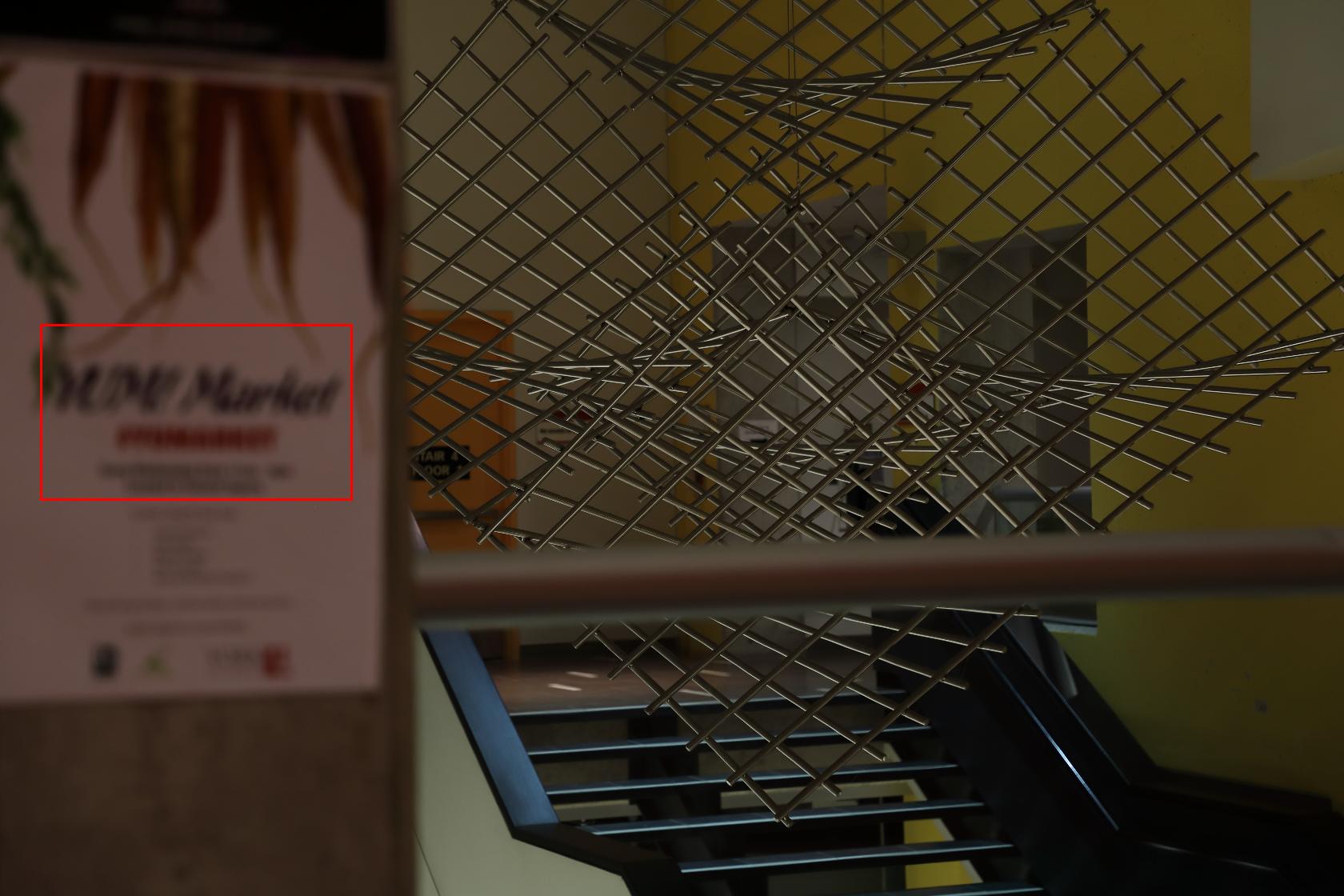}} &
                    \includegraphics[width=.145\textwidth]{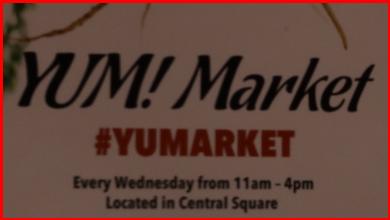} &
                    \includegraphics[width=.145\textwidth]{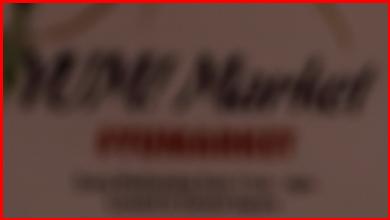} &
                    \includegraphics[width=.145\textwidth]{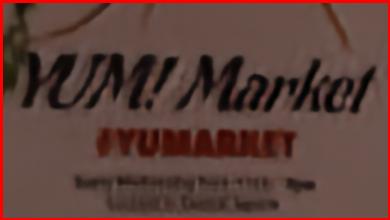} &
                    \includegraphics[width=.145\textwidth]{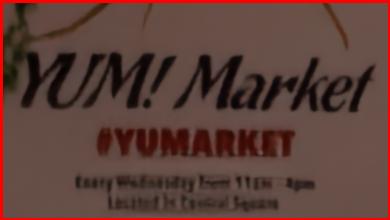}
                    \\
                    & \small~Reference & \small~Blurry & \small~DPDNet~\cite{abuolaim2020defocus} & \small~IFAN~\cite{lee2021iterative}
                    \\
                    &
                    \includegraphics[width=.145\textwidth]{images/dpdd/crop+Uformer} &
                    \includegraphics[width=.145\textwidth]{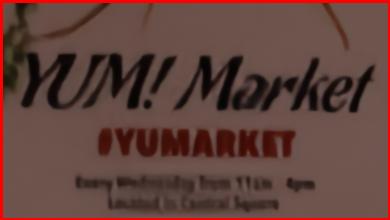} &
                    \includegraphics[width=.145\textwidth]{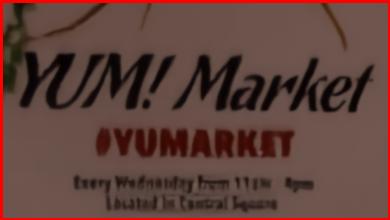} &
                    \includegraphics[width=.145\textwidth]{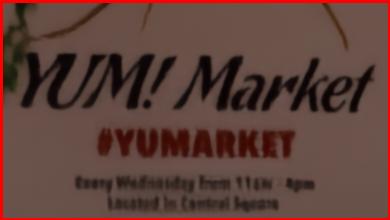}
                    \\
                    \small~Blurry Image & \small~Uformer~\cite{wang2021uformer} & \small~Restormer~\cite{Zamir2021Restormer} & \small~GRL~\cite{li2023grl} & \small~\textbf{Ours}
                \end{tabular}
            }
        \end{center}
        \caption{{Visual comparison on the DPDD dataset~\cite{abuolaim2020defocus}.}}
        \vspace{-1em}
    \end{figure*}

    \subsection{Motion Deblurring Results}
    Experimental results for motion deblurring are shown in Table~\ref{table:GoPro}.
    Our proposed method also achieved the leading performance.

    \begin{table}[h]
        \caption{Single-image motion deblurring results on the GoPro dataset~\cite{nah2017deep}.}
        \label{table:GoPro}
        \resizebox{\linewidth}{!}{
            \begin{tabular}{c|ccccccccc|c}
                \hline
                \multirow{2}{*}{Method} & MIMO-UNet & HINet & MAXIM & Restormer & UFormer & DeepRFT & MPRNet & NAFNet
                & GRL & Swintormer \\
                & ~\cite{cho2021rethinking} & ~\cite{chen2021hinet} & ~\cite{tu2022maxim} & ~\cite{Zamir2021Restormer} & ~\cite{wang2021uformer} & ~\cite{mao2021deep} & -local~\cite{
                    chu2021revisiting}  & ~\cite{chen2022simple}
                & ~\cite{li2023grl} & \textbf{ours} \\
                \hline
                PSNR & 32.68 & 32.71 & 32.86 & 32.92 & 32.97 & 33.23 & 33.31 & 33.69 & \textbf{33.93} & 33.38 \\
                SSIM & 0.959 & 0.959 & 0.961 & 0.961 & 0.967 & 0.963 & 0.964 & 0.967 & \textbf{0.968} & 0.965 \\
                \hline
            \end{tabular}
        }
    \end{table}

    \begin{figure*}
        \begin{center}
            \resizebox{\linewidth}{!}{
                \begin{tabular}[b]{c@{ } c@{ }  c@{ } c@{ } c@{ }   }
                    \hspace{-4mm}
                    \multirow{3}{*}[18pt]{\includegraphics[width=0.28\textwidth]{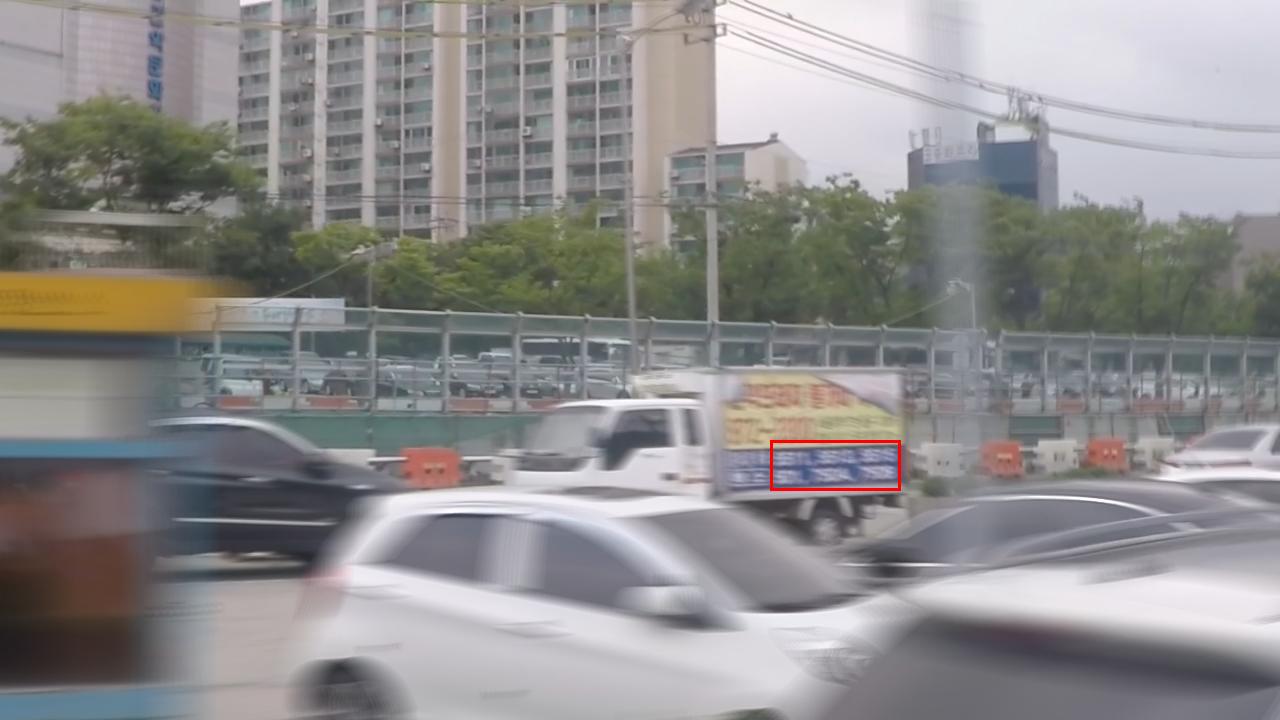}} &
                    \includegraphics[width=.145\textwidth]{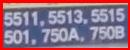} &
                    \includegraphics[width=.145\textwidth]{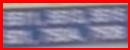} &
                    \includegraphics[width=.145\textwidth]{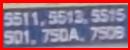} &
                    \includegraphics[width=.145\textwidth]{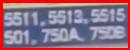}
                    \\
                    & \small~Reference & \small~Blurry & \small~SRN~\cite{tao2018srndeblur} & \small~MPRNet~\cite{Zamir2021MPRNet}
                    \\
                    &
                    \includegraphics[width=.145\textwidth]{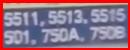} &
                    \includegraphics[width=.145\textwidth]{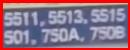} &
                    \includegraphics[width=.145\textwidth]{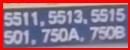} &
                    \includegraphics[width=.145\textwidth]{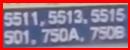}
                    \\
                    \small~Blurry Image & \small~Restormer~\cite{Zamir2021Restormer} & \small~NAFNet~\cite{chen2022simple} & \small~GRL~\cite{li2023grl} & \small~\textbf{Ours}
                \end{tabular}
            }
        \end{center}
        \caption{{Visual comparison on the GoPro dataset~\cite{nah2017deep}.}}
        \vspace{-1em}
    \end{figure*}

    \section{Ablation Studies} \label{Ablation}
    In this section, we investigate the effectiveness of the different designs of our proposed method.
    All experiments are conducted on the DPDD dataset.
    Previous findings have shown that our contributions have led to significant performance enhancements.
    Next, we will analyze the influence of each component individually.

    \begin{table}
        \caption{Ablation experiments. We train and test models on the DPDD dataset.
        For the baseline, we apply Restormer~\cite{Zamir2021Restormer}, a Transformer architecture based on channel attention(MDTA).
            $T$ represents the iteration numbers in the diffusion model.}
        \label{table:ablation_experiments}
        \resizebox{\linewidth}{!}{
            \begin{tabular}{c|c|c|c|c}
                \hline
                {Network} & Component
                & Params (M) & MACs (G) & PSNR (dB) \\
                \hline
                Baseline~\textbf{\textcolor{blue}{a}}                                  & Transformer(MDTA)                                             & 25.05  & 1.93  & 26.66 \\
                \hline
                \multirow{2}{*}{Transformer block~\textbf{\textcolor{blue}{b}}}        & Transformer(Swin) & 25.18 & 2.20 & 26.71 \\
                & Transformer(Swin+MDTA)                                        & 25.18  & 2.20  & 26.74 \\
                \hline
                \multirow{3}{*}{Pre-processing inference~\textbf{\textcolor{blue}{c}}} & window size(256)+shift size(220)+Transformer(Swin+MDTA) & 25.18  & 2.20 & 26.84  \\
                & window size(512)+shift size(220)+Transformer(Swin+MDTA)       & 25.18  & 2.20  & 26.98 \\
                & window size(512)+shift size(220)+Transformer(MDTA)            & 26.13  & 18.70 & 26.91 \\
                & window size(512)+shift size(384)+Transformer(Swin+MDTA)       & 25.18  & 2.20  & 26.89 \\
                \hline
                \multirow{4}{*}{Diffusion prior~\textbf{\textcolor{blue}{d}}}          & T(5)+Transformer(MDTA) & 138.8 & 8.02 & 26.67 \\
                & T(10)+Transformer(MDTA)                                       & 138.8  & 8.02  & 26.71 \\
                & T(20)+Transformer(MDTA)                                       & 138.8  & 8.02  & 26.75 \\
                & T(50)+Transformer(MDTA)                                       & 138.8  & 8.02  & 26.77 \\
                \hline
                Overall                                                                & T(50)+window size(512)+shift size(220)+Transformer(Swin+MDTA) & 154.89 & 8.02 & 27.07  \\
                \hline
            \end{tabular}
        }
    \end{table}

    \noindent \textbf{Improvements in mixed attention.}
    Table~\ref{table:ablation_experiments}{\textcolor{blue}{b}} shows that the Shifted Windows-Dconv Attention has comparable performance with MDTA\@.
    Furthermore, introducing the Shifted Windows-Dconv Attention to MDTA brings a better performance.
    Overall, our proposed Transformer block contributions lead to a gain of 0.08 dB over the baseline.

    \noindent \textbf{Improvements in pre-processing inference.}
    Table~\ref{table:ablation_experiments}{\textcolor{blue}{c}} shows that the pre-processing inference has comparable performance with TLC~\cite{chu2021tlc}.
    Furthermore, it’s noteworthy that a significant performance improvement of 0.25 dB over the baseline has been achieved by simply adjusting the window size and shift size to align the input tensor size with the training tensor size, without requiring retraining or fine-tuning.

    \noindent \textbf{Impact of diffusion prior.}
    We construct a baseline model without priors generated by diffusion model.
    Table~\ref{table:ablation_experiments}{\textcolor{blue}{d}} demonstrates that the diffusion priors provide favorable gain of 0.11dB over the baseline.
    Furthermore, we explore the impact of the iteration numbers $T$ in the diffusion model.
    A larger number of iterations leads the diffusion model to generate more accurate features.
    Therefore, the corresponding deblurring model can utilize the features more accurately.
    Based on testing results, when the number of iterations reaches 20, the improvement of the entire deblurring model gradually converges.
    However, for better performance, we choose $T=50$ for the final model.

    \section{Conclusion}
    We endeavor to extend the applicability of deep learning deblurring methods beyond laboratory settings, aiming to achieve favorable results across a broader spectrum of real-world scenarios.
    Our principal contributions are presenting a new model and a new inference strategy that make deblur high-resolution images in personal computer possible.
    Specifically, integrated with DM, a memory-efficient model named Swintormer was built.
    It is an image deblurring Transformer model, designed to efficiently process high-resolution images with remarkably low MACs.
    The proposed Transformer block demonstrates improved performance by applying self-attention mechanisms across both channel and spatial dimensions, while maintaining linear complexity.
    Furthermore, we present a plug and play methodology that ensures consistency in input tensor sizes during training and inference, thereby enhancing model performance.
    Importantly, this approach obviates the need for retraining or fine-tuning, resulting in performance enhancements across various tasks.

    \bibliographystyle{splncs04}
    \bibliography{main}

\end{document}